\documentclass[letterpaper, 10pt, conference]{ieeeconf}

\IEEEoverridecommandlockouts  
\usepackage[english]{babel}
\usepackage[utf8]{inputenc}
\usepackage{times} 
\usepackage{cite}

\usepackage{booktabs}

\usepackage[hidelinks]{hyperref}
\usepackage[nolist]{acronym}

\usepackage{xurl}

\usepackage{multirow}

\usepackage{subcaption}

\usepackage{algorithm}
\usepackage{algorithmic}

\usepackage{siunitx}
\usepackage{graphicx} 
\usepackage{epsfig} 
\usepackage{pgfplots}
\pgfplotsset{compat=1.18}
\pgfplotsset{every axis/.append style={grid=major, grid style={gray!25, line width=0.3pt}}}
\usepgfplotslibrary{statistics}
\usepgfplotslibrary{groupplots}

\usepackage{float}

\usepackage[export]{adjustbox}

\usepackage{xcolor}
\definecolor{myYellow}{rgb}{0.93,0.69,0.13}
\definecolor{myPurple}{rgb}{0.49,0.18,0.56}
\definecolor{myGreen}{rgb}{0.26 0.72 0.54}
\definecolor{darkgreen}{rgb}{0.272, 0.50, 0.376}
\definecolor{lightgreen}{rgb}{0.585, 0.82, 0.647}

\definecolor{cOrange}{RGB}{255,127,0}
\definecolor{cRed}{RGB}{214,39,40}
\definecolor{cBlue}{RGB}{31,119,180}
\definecolor{cGrey}{RGB}{127,127,127}
\definecolor{cGreen}{RGB}{44,160,44}

\colorlet{mydarkblue}{blue!30!black}

\usepackage{tikz}
\pgfdeclarelayer{foreground}
\pgfsetlayers{background, main, foreground}
\usetikzlibrary{quotes, angles, backgrounds, arrows, automata, shapes, positioning, calc, through, spy, decorations.pathreplacing, decorations.markings, arrows.meta, automata, petri, shapes.multipart, patterns}

\newlength{\panelwd}
\newlength{\panelht}

\tikzset{
    imglabel/.style={
      rectangle,
      inner sep=2pt,
      text=black,
      minimum height=1em,
      text centered,
      fill=white,
      fill opacity=1.0,
      text opacity=1,
      anchor=south west,
    },
  }
\tikzset{
	state/.style={
		rectangle,
		draw=black, very thick,
		minimum height=1.0em,
		text centered,
	},
}
\tikzset{
  on each segment/.style={
    decorate,
    decoration={
      show path construction,
      moveto code={},
      lineto code={
        \path [#1]
        (\tikzinputsegmentfirst) -- (\tikzinputsegmentlast);
      },
      curveto code={
        \path [#1] (\tikzinputsegmentfirst)
        .. controls
        (\tikzinputsegmentsupporta) and (\tikzinputsegmentsupportb)
        ..
        (\tikzinputsegmentlast);
      },
      closepath code={
        \path [#1]
        (\tikzinputsegmentfirst) -- (\tikzinputsegmentlast);
      },
    },
  },
  mid arrow/.style={postaction={decorate,decoration={
        markings,
        mark=at position .5 with {\arrow[#1]{stealth}}
      }}},
}

\tikzset{
  half circle/.style={
      semicircle,
      shape border rotate=180,
      anchor=chord center,
      minimum size=5mm
      }
}

\usepackage{mathtools}
\usepackage{nicefrac}
\usepackage{amsmath}
\usepackage{amssymb}
\usepackage{bm} 
\usepackage{scalerel} 

\usepackage{amsthm}

\newtheoremstyle{compact}
  {4pt}   
  {4pt}   
  {\itshape}
  {}
  {\bfseries}
  {.}
  {0.5em}
  {}

\DeclareMathAlphabet{\pazocal}{OMS}{zplm}{m}{n}

\usepackage{amsthm}
\newtheoremstyle{compact}
  {4pt}   
  {4pt}   
  {\itshape}
  {}
  {\bfseries}
  {.}
  {0.5em}
  {}

\theoremstyle{compact}
\newtheorem{theorem}{Theorem}
\newtheorem{lemma}[theorem]{Lemma}
\newtheorem{proposition}[theorem]{Proposition}
\newtheorem{corollary}[theorem]{Corollary}
\newtheorem{assumption}{Assumption}

\newtheorem{problem}{Problem}

\newcommand{\vv}{\mathbf{v}}                      
\newcommand{\wv}{\mathbf{w}}                      
\newcommand{\tauv}{\boldsymbol{\tau}}             
\newcommand{\taubar}{\bar{\tau}}                  
\newcommand{\dv}{\mathbf{f}}                      
\newcommand{\Dm}{\mathbf{D}}                      
\newcommand{\Am}{\mathbf{A}}                      
\newcommand{\Mm}{\mathbf{M}}                      
\newcommand{\Jm}{\mathbf{J}}                      
\newcommand{\lbar}{\bar{\ell}}                    
\newcommand{\ldrop}{\ell^{\dagger}}               
\newcommand{\Lop}{L^{\mathrm{op}}}                
\newcommand{\Lmax}{L^{\max}}                      
\newcommand{\Cset}{\mathcal{C}}                   
\newcommand{\vsat}{v^{\mathrm{sat}}}              
\newcommand{\vstar}{v^{\star}}                    
\newcommand{\Bset}{\mathcal{B}}                   
\newcommand{\vlo}{\underline{\mathbf{v}}}         
\newcommand{\vhi}{\overline{\mathbf{v}}}          
\newcommand{\psilo}{\underline{\psi}}             
\newcommand{\Lbox}{\underline{L}}                 
\newcommand{\taudlo}{\underline{d}}               
\newcommand{\taudhi}{\overline{d}}                
\newcommand{\taudnom}{\hat{d}}                    
\newcommand{\jit}{\Delta d}                       
\newcommand{\tauh}{\boldsymbol{\tau}^{\mathrm{h}}}
\newcommand{\Iset}{\mathcal{I}}                   
\newcommand{\cert}{\mathcal{L}_k}                 
\newcommand{\epsm}{\varepsilon}                   
\newcommand{\veq}{v^{\mathrm{eq}}}                

\graphicspath{{./figures/}}

\title{\LARGE \bf Delay-Certified Readiness Barriers for Overactuated\\ Multirotor Allocation Without Lipschitz Constants}

\author{Giuseppe Silano,~\IEEEmembership{Senior Member,~IEEE}
    \thanks{$^1$Giuseppe Silano is 
    with the Multi-robot Systems (MRS) group, Department of Cybernetics, Czech Technical University, 12135 Prague, Czech Republic (e-mail: {\tt\small giuseppe.silano@fel.cvut.cz}).}
    \thanks{This work was partially funded by 
    the CTU grant no. SGS26/077/OHK3/1T/13 and the GAČR project no.~27-00437M.}
    \thanks{The authors used Google Gemini 4 and Anthropic Claude Opus 5.5 for proofreading and exposition refinement, but independently verified all outputs and assumes full responsibility for the article's content.}
}

\begin{document}

\maketitle
\thispagestyle{empty} 
\pagestyle{empty} 


\begin{acronym}
    \acro{CBF}[CBF]{Control Barrier Function}
    \acro{ESC}[ESC]{Electronic Speed Controller}
    \acro{QP}[QP]{Quadratic Program}
    \acro{RMS}[RMS]{Root Mean Square}
\end{acronym}



\begin{abstract}
    Overactuated multirotors can produce the same wrench with many rotor-speed combinations, some of which leave the vehicle unable to change its wrench quickly, e.g., to reject a gust. Readiness barriers keep this ability, the readiness, above a floor, but only in continuous time; with sampled commands and varying speed-controller delays, the available robustness bound covers less than one millisecond. We remove this limit by exploiting the structure of the problem: each rotor responds monotonically and in closed form to its own torque, so the speeds reachable under uncertain switching times lie in a cheaply computed box, and the lowest readiness over that box sits at a corner known in advance. This yields a Lipschitz-free test that a command keeps the readiness above its floor until the next sample, plus a fallback torque that keeps the vehicle on a certified plan. In hexarotor simulations with 2--40~ms delays and $\pm2$~ms jitter, the filter never violated the floor. At $10\pm2$~ms, it matched a delay-compensating predictor within the design envelope and cost 2--30\% more tracking error beyond it, where the predictor alone violated the floor in 2 to 16 of 16 missions. A tuned robust predictor was also safe with synchronous switching, but not with per-rotor jitter.
\end{abstract}






\section{Introduction}
\label{sec:intro}

Overactuated multirotors carry more rotors than the components of the wrench (the force and moment acting on the body) they control, which enables full-pose tracking and physical interaction~\cite{ryll2015overactuated, hamandi2021taxonomy}. A continuum of rotor speeds then produces the commanded wrench, and the allocator selects one, typically by minimizing effort subject to actuator limits~\cite{johansen2013allocation, ciresola2026review}. Effort minimization, however, is blind to the authority to \emph{change} the wrench. Thrust and drag scale quadratically with spin rate~\cite{mahony2012multirotor}, so a slow rotor has a shallow thrust slope and a fast one has spent its torque on drag; either way, the vehicle cannot respond quickly to a disturbance.

The drag-aware co-metric of~\cite{franchi2026promptness} makes this authority measurable. A rotor changes its thrust at a rate given by its thrust slope, which grows with speed, times the acceleration it can still produce in either direction, which shrinks as drag absorbs its torque. The co-metric maps these rates through the allocation matrix to the set of achievable wrench rates, whose log-volume we call the \emph{readiness}, in the spirit of manipulability~\cite{yoshikawa1985manipulability}: it is low when a rotor is nearly stopped or saturated, and $-\infty$ when some wrench component can no longer change. Rather than maximizing it, which leads to discontinuous allocations, our previous work~\cite{silano2026readiness} keeps it above a floor with a~\ac{CBF}~\cite{ames2017cbfqp, ames2019ecc} enforced by a torque-level~\ac{QP}. This \emph{readiness barrier} certifies the floor, but, like most \ac{CBF} guarantees, in continuous time; the present paper extends it to sampled, delayed implementations.

Flight controllers, however, hold each command for a sampling period and send it through \acp{ESC} whose delay varies from sample to sample. Between two samples the barrier condition is not enforced, and with delay the filter acts on a past state. In~\cite{silano2026readiness}, simulations of a hexarotor showed that the filter tolerates \ac{ESC} delays of 1--10~ms, but its robustness bound~\cite[Prop.~13]{silano2026readiness}, built on Lipschitz constants of the barrier that are worst cases over the whole certified set, certifies only 0.63~ms there: the guarantee lags behind the observed robustness. At longer delays, or with more demanding maneuvers, it can violate the floor: at 40~ms on a nominal mission, and from 5~ms when the maneuver demands more than the floor was designed for (Section~\ref{sec:sim}). This paper closes the gap between what is observed and what is certified.

Our key observation is that the readiness barrier has enough structure to be checked exactly between samples, without Lipschitz constants. Each rotor responds to its own torque independently, monotonically, and in closed form, so the speeds reachable under any admissible switching times lie in a box whose corners are computed directly. Moreover, since a slow rotor lacks thrust slope and a fast one lacks torque margin, the contribution of each rotor to the readiness first grows and then shrinks with its speed (Lemma~\ref{lem:unimodal} and Fig.~\ref{fig:concept}(c)), so the lowest readiness over the box sits at a corner known in advance. A command is then accepted only if the readiness provably stays above the floor until the next sample. Our contributions are:
\begin{enumerate}
    \item An exact, closed-form minimum of the readiness over a box of rotor speeds (Proposition~\ref{prop:boxmin}).
    \item A delay certificate for bounded, order-preserving, per-\ac{ESC} delays, with a hold torque that keeps the last certified plan safe and makes the test recursively feasible (Theorem~\ref{thm:delay}), and a bound tracing its conservatism to the jitter and the command variation (Corollary~\ref{cor:jitter}).
    \item An extension to bounded drag and speed-measurement errors (Proposition~\ref{prop:robust}).
    \item A certified allocation filter (Algorithm~\ref{alg:filter}), compared on the hexarotor of~\cite{silano2026readiness} with predictor-based and Lipschitz-margin filters (Section~\ref{sec:sim}).
\end{enumerate}



\section{Related Work}\label{sec:related}

\emph{Allocation and actuator authority.} Allocation is classically a constrained least-squares or linear program~\cite{johansen2013allocation}; \cite{hamandi2021taxonomy, ciresola2026review} survey multirotor and tilting-platform allocation. These methods make the wrench attainable but do not preserve the authority to change it. Barriers on actuator quantities appear in \ac{CBF}-based allocation for obstacle avoidance~\cite{yang2024allocation} and in servo-rate allocation for tiltrotors~\cite{lee2026tiltrotor}; \cite{silano2026readiness} and~\cite{silano2026tilt} floor the readiness with fixed and articulated rotors, and matrix \acp{CBF}~\cite{ong2026mcbf} offer a directional alternative to the log-determinant.

\emph{\acp{CBF} under sampling and delay.} \ac{CBF} filters~\cite{ames2017cbfqp, ames2019ecc} are designed in continuous time; for the readiness barrier, which has relative degree one, the issue is their sampled, delayed implementation. Sampled-data methods tighten the barrier condition with margins from global or local bounds~\cite{breeden2022sampled}, use approximate discrete-time models~\cite{taylor2022sampled}, or certify a given controller over the hold interval~\cite{bahati2024sample}; the local conditions of~\cite{breeden2022sampled} need no Lipschitz constant but a supremum over the reachable set. For input delay, \acp{CBF} act on a predicted state~\cite{jankovic2018delay, molnar2023delay}, made robust to prediction errors by input-to-state safety~\cite{molnar2022issf}, robust formulations~\cite{molnar2023delay}, delay estimation~\cite{kim2024delayest}, or integral \acp{CBF}~\cite{kiss2026icbf}; time-varying delays are treated in~\cite{abel2021tvdelay}, and interval analysis handles measurement and actuation uncertainty~\cite{zhang2022interval}.

\emph{Reachability and backup controllers.} Monotone and mixed-monotone systems give cheap interval over-approximations of reachable sets~\cite{angeli2003monotone, coogan2020mixed}. Backup \acp{CBF} integrate a backup controller over a horizon~\cite{chen2021backup}, also under input bounds~\cite{vanwijk2026backup} and mixed state-input constraints~\cite{gacsi2026multiple}; \cite{singletary2020sampled} adds sampled data, state uncertainty, and a known delay, and~\cite{abate2020runtime} minimizes a concave barrier over all $2^n$ vertices of mixed-monotone backup boxes. 

\emph{Positioning.} We specialize these ingredients: for general dynamics, the reachable set and the extremum of the barrier over it must be approximated, whereas for readiness barriers the set is a box computed in closed form, the nonconcave barrier is minimized exactly at one known vertex, and the infinite tail of the backup needs no integration.



\section{Problem Formulation}\label{sec:prelim}

This section recalls the readiness barrier of~\cite{silano2026readiness} and shows why sampling and delay break its guarantee.

\subsection{Model and readiness barrier}
We use the model and notation of~\cite{silano2026readiness}. An overactuated multirotor with $n$ rotors produces the wrench $\wv=\Am\boldsymbol{\phi}(\vv)\in\mathbb{R}^m$, $n>m$, with $\boldsymbol{\phi}(\vv)=\vv\odot|\vv|$ (componentwise $\phi(v)=v|v|$), and the spin rate $v_i$ of rotor $i$ obeys
\begin{equation}\label{eq:rotor}
    m_i\dot v_i=-b_i v_i|v_i|+\tau_i,\qquad |\tau_i|\le\taubar_i ,
\end{equation}
where $m_i>0$ is the inertia of motor and propeller, $b_i>0$ the aerodynamic drag coefficient, and $\tau_i$ the motor torque, bounded by $\taubar_i>0$; in vector form, $\dot{\vv}=\dv(\vv)+\Mm^{-1}\tauv$ with $\Mm=\mathrm{diag}(m_i)$, and $\mathrm{rank}\,\Am=m$. The motor time scale is $T_0=m_i/\sqrt{\taubar_ib_i}$. The symmetric acceleration capacity $\bar a_i(v_i)=(\taubar_i-b_iv_i^2)/m_i$ is the acceleration rotor $i$ can still produce in either direction; it is positive on the open box $\mathcal{V}=\{\vv:|v_i|<\vsat_i\}$, where $\vsat_i:=\sqrt{\taubar_i/b_i}$ is the saturation speed, at which drag absorbs the full torque, and admissible torques render $\mathcal{V}$ forward invariant; and on $\mathcal{V}$ the drag-aware co-metric and readiness level are~\cite{franchi2026promptness, silano2026readiness}
\begin{equation}\label{eq:readiness}
    \begin{gathered}
    \Dm(\vv)=4\Am\boldsymbol{\Psi}(\vv)\Am^{\top},\quad
    \psi_i(v_i)=v_i^2\bar a_i(v_i)^2,\\
    L(\vv)=\ln\det\Dm(\vv),
    \end{gathered}
\end{equation}
with $\boldsymbol{\Psi}=\mathrm{diag}(\psi_i)$ and $\Am_{\bullet i}$ the $i$th column of $\Am$. The weight $\psi_i$ is a quarter of the squared largest rate, $2v_i\bar a_i$, at which rotor $i$ can change its thrust $\phi(v_i)$ in either direction, so $\Dm=\Jm\,\mathrm{diag}(\bar a_i^2)\Jm^{\top}$, with $\Jm$ defined below, is the shape matrix of the ellipsoid $\{\Jm\,\mathrm{diag}(\bar a_i)\mathbf{u}:\|\mathbf{u}\|\le1\}$ of wrench rates available at $\vv$, and the readiness $L$ is, up to a constant, twice the logarithm of its volume. Each $\psi_i$ vanishes at standstill (no thrust slope) and at saturation (no torque margin) and peaks at the sweet spot $\vstar_i=\vsat_i/\sqrt3$ (Lemma~\ref{lem:unimodal}). Given a floor $\lbar$, the barrier is $h(\vv)=L(\vv)-\lbar$ and the certified set is $\Cset=\{\vv\in\mathcal{V}:h(\vv)\ge 0\}$. The floor is chosen in the window $\ldrop<\lbar\le\Lop$. The lower end is the readiness left when the most dispensable rotor stops, so a floor above it keeps every rotor spinning; the upper end is the best readiness that exact tracking guarantees throughout the mission, so a floor below it does not force tracking errors. Formally, $\ldrop=\Lmax-\min_k\Delta_k$, where $\Delta_k=-\ln(1-\lambda_k)$ is the readiness lost when rotor $k$ stops, with leverage $\lambda_k=\psi_k(\vstar_k)\Am_{\bullet k}^{\top}\mathbf{S}^{-1}\Am_{\bullet k}$, $\mathbf{S}=\sum_i\psi_i(\vstar_i)\Am_{\bullet i}\Am_{\bullet i}^{\top}$, $\Lmax=\ln\det(4\mathbf{S})$,
and $\Lop=\min_{\wv\in\mathcal{W}}\max_{\vv:\Am\boldsymbol{\phi}(\vv)=\wv}L(\vv)$ is the worst-case achievable readiness over the mission wrenches $\mathcal{W}$~\cite[Thm.~1, eq.~(10)]{silano2026readiness}. Because $\lbar>\ldrop$, trajectories in $\Cset$ keep the sign of every rotor speed, which by symmetry of \eqref{eq:rotor} we take positive.

The filter of~\cite{silano2026readiness} solves, at the measured state, the \ac{QP}
\begin{equation}\label{eq:qp}
    \begin{aligned}
    \min_{\tauv,\boldsymbol{\delta}}\;& \|\tauv-\tauv_{\mathrm{ref}}\|^2+\rho\|\boldsymbol{\delta}\|^2\\
    \text{s.t.}\;& \nabla h(\vv)^{\top}\mathbf{a}\ge-\alpha h(\vv),\quad
    \Jm(\vv)\mathbf{a}=\boldsymbol{\mu}+\boldsymbol{\delta},\\
    & |\tauv|\le\bar{\tauv},\quad \mathbf{a}=\dv(\vv)+\Mm^{-1}\tauv,
    \end{aligned}
\end{equation}
where $\Jm=2\Am\,\mathrm{diag}(|\vv|)$, $\boldsymbol{\mu}=\dot{\wv}_{\mathrm{des}}+k_w(\wv_{\mathrm{des}}-\Am\boldsymbol{\phi}(\vv))$ is the task demand for the desired wrench $\wv_{\mathrm{des}}$, $\tauv_{\mathrm{ref}}$ is a nominal allocation (in~\cite{silano2026readiness} and here, the solution of \eqref{eq:qp} without the barrier row and with $\tauv_{\mathrm{ref}}=\mathbf{0}$, i.e., the effort-minimizing allocation), and $\alpha,k_w,\rho>0$ are gains. In words, the \ac{QP} stays close to the nominal allocation, tracks the wrench demand up to the penalized slack $\boldsymbol{\delta}$, and lets $h$ decrease at most at rate $\alpha h$, so the floor has priority over tracking. Applied in continuous time, \eqref{eq:qp} renders $\Cset$ forward invariant~\cite[Thm.~7]{silano2026readiness}.



\subsection{Sampled data and actuation delay}
On a flight controller, \eqref{eq:qp} is solved at $t_j=j\Delta t$ and its output passes through the \acp{ESC} with a delay that varies from sample to sample. We model this as follows.

\begin{assumption}\label{ass:delay}
The command $\tauv_j$ (components $\tau_{j,i}$), computed at $t_j$ from the exact state $\vv(t_j)$, is applied to rotor $i$ on $[\sigma_{j,i},\sigma_{j+1,i})$, where the unknown switching instants satisfy $\sigma_{j,i}\in[t_j+\taudlo,\,t_j+\taudhi]$, $0\le\taudlo\le\taudhi$, and $\sigma_{j,i}\le\sigma_{j+1,i}$ (commands are not reordered; each \ac{ESC} may switch at its own instant). The commands issued before $t=0$ are known. The jitter width is $\jit:=\taudhi-\taudlo$.
\end{assumption}

In words, each command reaches each \ac{ESC} after an unknown delay between $\taudlo$ and $\taudhi$, the \acp{ESC} may switch at different instants, and no command overtakes an earlier one. The lower bound $\taudlo$ includes the computation time of the filter. For a filter that ignores the delay, \cite[Prop.~13, Cor.~14]{silano2026readiness} absorbs a delay $d$ by raising the floor by $\eta(d)=(\alpha\bar H+KVd)d+KVd/\alpha$, where $\bar H=\Lmax-\lbar$ and $K$, $V$ bound the second derivatives of $h$ and the rotor accelerations on $\Cset$; keeping the raised floor below $\Lop$ caps the certified delay at $d^{\star}=0.63$~ms for the hexarotor of~\cite{silano2026readiness}. A predictor removes the nominal delay, but its error under jitter again calls for a margin. We therefore pose the following problem.

\begin{problem}\label{prob:cert}
Under Assumption~\ref{ass:delay}, given $\vv_k$ and the committed commands $\{\tauv_j\}_{j<k}$, decide without Lipschitz constants whether a candidate $\tauv_k$ guarantees $h(\vv(t))\ge0$ on $[t_k,t_{k+1}]$ for every admissible realization of the switching instants, and ensure that some candidate always passes.
\end{problem}



\section{A Lipschitz-Free Delay Certificate}\label{sec:cert}

This section solves Problem~\ref{prob:cert}. The solution rests on three facts: every rotor is a scalar, monotone system with a closed-form flow; each $\psi_i$ is unimodal; and $L$ is monotone in every $\psi_i$.



\subsection{Three facts}

\begin{lemma}[Monotone rotor flow]\label{lem:flow}
Under a constant torque $\tau_i$, the solution $t\mapsto v_i(t)$ of \eqref{eq:rotor} is monotone in $t$, and $v_i(t)$ is nondecreasing in $v_i(0)$ and in $\tau_i$. For $v_i(0)=v_0\ge 0$, let $v^{\infty}=\sqrt{|\tau_i|/b_i}$ and $\omega=\sqrt{|\tau_i|b_i}/m_i$. If $\tau_i>0$, $v_i(t)=v^{\infty}\tanh(\omega t+\mathrm{artanh}(v_0/v^{\infty}))$ for $v_0<v^{\infty}$, with $\coth$ and $\mathrm{arcoth}$ for $v_0>v^{\infty}$; if $\tau_i=0$, $v_i(t)=v_0/(1+b_iv_0t/m_i)$; if $\tau_i<0$, $v_i(t)=v^{\infty}\tan(\arctan(v_0/v^{\infty})-\omega t)$ up to the zero crossing $t_z$, and $-v^{\infty}\tanh(\omega(t-t_z))$ after it. For $v_0<0$, $v_i(t;v_0,\tau_i)=-v_i(t;-v_0,-\tau_i)$.
\end{lemma}
\begin{proof}
The right-hand side of \eqref{eq:rotor} is locally Lipschitz in $v_i$ and nondecreasing in $\tau_i$. Solutions of a scalar autonomous equation are monotone in time, and order is preserved in the initial condition and in the input by the comparison principle for scalar (hence monotone) systems~\cite{angeli2003monotone}. The closed forms follow by substitution into \eqref{eq:rotor}.
\end{proof}

Lemma~\ref{lem:flow} makes every rotor speed under a piecewise-constant torque computable in closed form, and ordered: more torque or a faster start never yields a slower rotor. 

\begin{lemma}[Unimodality]\label{lem:unimodal}
On $[0,\vsat_i]$, $\psi_i$ is increasing on $[0,\vstar_i]$ and decreasing on $[\vstar_i,\vsat_i]$, with $\vstar_i=\vsat_i/\sqrt 3$. Hence, for $0\le l\le u\le\vsat_i$, $\min_{v\in[l,u]}\psi_i(v)=\min\{\psi_i(l),\psi_i(u)\}$.
\end{lemma}
\begin{proof}
$\psi_i'(v)=2v\bar a_i(v)(\taubar_i-3b_iv^2)/m_i$~\cite[Lemma~3]{silano2026readiness}, positive on $(0,\vstar_i)$ and negative on $(\vstar_i,\vsat_i)$, so the minimum over an interval is at an endpoint.
\end{proof}

For a box $\Bset=[\vlo,\vhi]$ of spin rates with $\vhi\le\vsat$, define
\begin{equation}\label{eq:boxmin}
    \psilo_i(\Bset)=\begin{cases}\min\{\psi_i(\underline v_i),\psi_i(\overline v_i)\}, & \underline v_i>0,\\ 0, & \underline v_i\le 0,\end{cases}
\end{equation}
and $\Lbox(\Bset)=\ln\det\big(4\Am\,\mathrm{diag}(\psilo(\Bset))\Am^{\top}\big)$, with $\ln\det=-\infty$ for a singular matrix.

\begin{proposition}[Exact box minimum]\label{prop:boxmin}
With $\psilo$ as in \eqref{eq:boxmin}, if $\vlo>\mathbf{0}$ and $\vhi\le\vsat$, then $\min_{\vv\in\Bset}L(\vv)=\Lbox(\Bset)$. If $\underline v_i\le 0$ for some $i$, then $\Lbox(\Bset)\le\ldrop<\lbar$.
\end{proposition}
\begin{proof}
For $\vv\in\Bset$, Lemma~\ref{lem:unimodal} gives $\psi_i(v_i)\ge\psilo_i(\Bset)$ for every $i$, so, by \eqref{eq:readiness}, $\Dm(\vv)\succeq 4\Am\,\mathrm{diag}(\psilo)\Am^{\top}$ and $L(\vv)\ge\Lbox(\Bset)$
by monotonicity of $\ln\det$ on the positive semidefinite cone. Equality holds at the vertex attaining every $\psilo_i$. If $\psilo_i=0$, then $\psilo_j\le\psi_j(\vstar_j)$ for $j\ne i$ gives $4\Am\,\mathrm{diag}(\psilo)\Am^{\top}\preceq4(\mathbf{S}-\psi_i(\vstar_i)\Am_{\bullet i}\Am_{\bullet i}^{\top})$, and the matrix determinant lemma yields $\Lbox\le\Lmax+\ln(1-\lambda_i)=\Lmax-\Delta_i\le\ldrop$~\cite[Thm.~1]{silano2026readiness}.
\end{proof}

In words, the lowest readiness over a box is at the corner where each rotor sits at the end of its range with the smaller weight; finding it costs $2n$ scalar evaluations and one $m\times m$ log-determinant instead of a search over $2^n$ corners. Any conservatism thus lies in the box, not in the minimization.



\subsection{Reachable boxes under jittered delay}
Fix a step $k$. Commands $\tauv_j$ with $j<k$ are already committed, $\tauv_k$ is the candidate, and for $j>k$ we set $\tauv_j=\tauh$, the \emph{sweet-spot hold torque}
\begin{equation}\label{eq:hold}
    \tau^{\mathrm{h}}_i=b_i(\vstar_i)^2=\taubar_i/3,
\end{equation}
which makes $\vstar$ an equilibrium of \eqref{eq:rotor}. Under $\tauh$, every rotor moves toward the speed of maximum readiness, so readiness can only grow and nothing after the window needs checking. This hypothesis on future commands only decides whether $\tauv_k$ is accepted; it plays no role in the guarantee over the current interval (Theorem~\ref{thm:delay}(i)).

Let $\Iset(t)=\{j:t_j+\taudlo\le t<t_{j+1}+\taudhi\}$ and define the torque envelopes $\tau^{+}_i(t)=\max_{j\in\Iset(t)}\tau_{j,i}$ and $\tau^{-}_i(t)=\min_{j\in\Iset(t)}\tau_{j,i}$. Let $\vv^{\pm}$ be the solutions of \eqref{eq:rotor} from $\vv^{\pm}(t_k)=\vv_k:=\vv(t_k)$ under $\tauv^{\pm}$.

\begin{lemma}[Envelopes]\label{lem:env}
Under Assumption~\ref{ass:delay}, the torque applied to rotor $i$ at time $t$ is $\tau_{j,i}$ for some $j\in\Iset(t)$, and $\vv^{-}(t)\le\vv(t)\le\vv^{+}(t)$ for all $t\ge t_k$ and every admissible realization of $\{\sigma_{j,i}\}$.
\end{lemma}
\begin{proof}
If $\tau_{j,i}$ is applied to rotor $i$ at $t$, then $\sigma_{j,i}\le t<\sigma_{j+1,i}$, so $t_j+\taudlo\le t$ and $t<t_{j+1}+\taudhi$; hence $j\in\Iset(t)$ and $\tau^{-}_i(t)\le\tau_i(t)\le\tau^{+}_i(t)$ for every $i$. The rotors are decoupled and each is monotone in its input (Lemma~\ref{lem:flow}), so the comparison principle~\cite{angeli2003monotone} gives the ordering.
\end{proof}

\begin{figure}[tb]
    \centering
    \resizebox{\columnwidth}{!}{\input{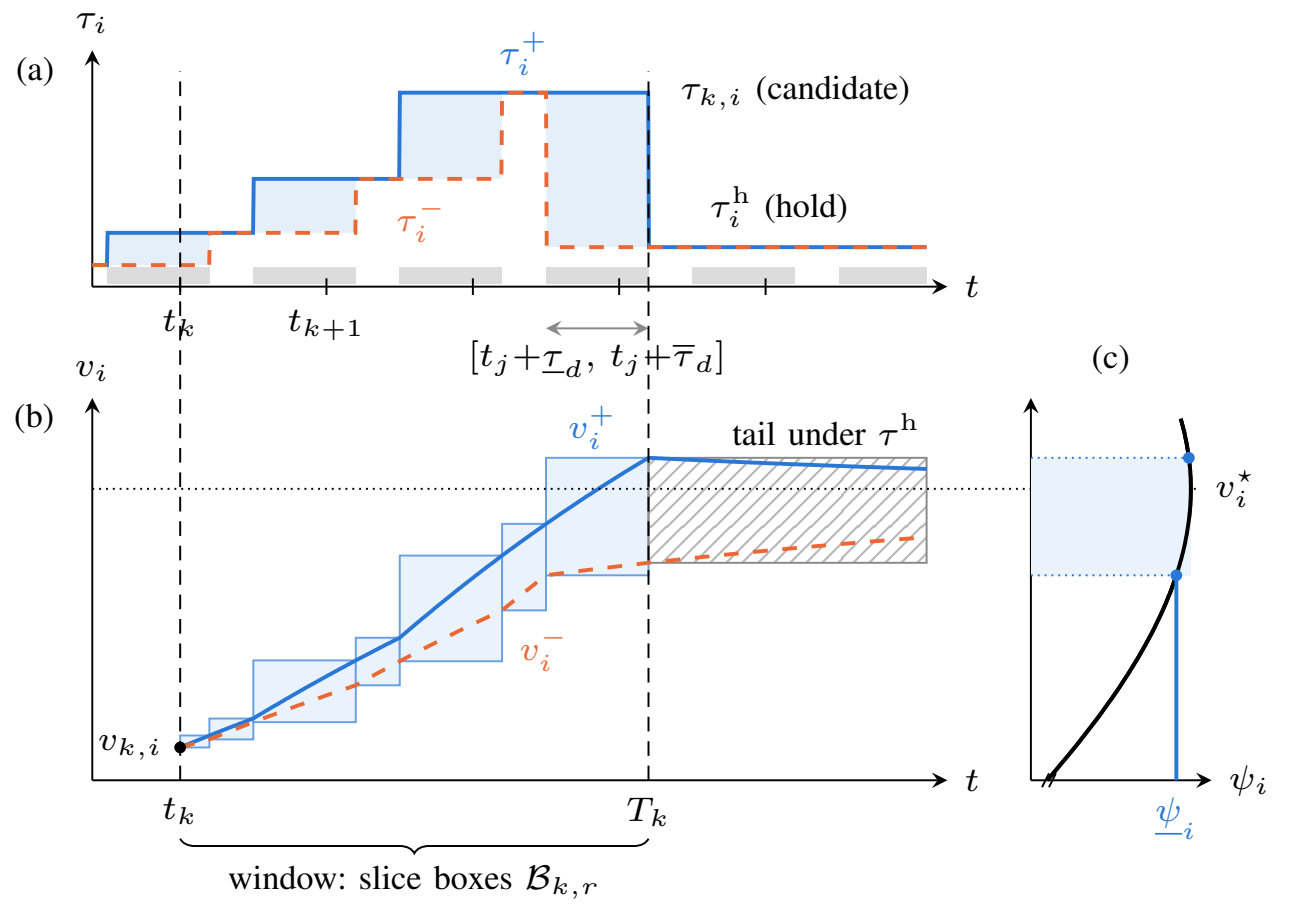}}
    \caption{Certificate of one rotor at step $k$ (illustrative timing). (a) Commands switch at unknown instants in the gray intervals; $\tau_i^{\pm}$ take the extreme command. (b) The envelopes $v_i^{\pm}$ and slice boxes $\mathcal{B}_{k,r}$ enclose every admissible trajectory; after $T_k$ the hold torque drives both toward $v_i^{\star}$. (c) $\psi_i$ is unimodal, so its minimum over a box side is at an endpoint.}
    \label{fig:concept}
\end{figure}

The envelopes (Figure~\ref{fig:concept}) are piecewise constant, with breakpoints in $\{t_j+\taudlo,\,t_j+\taudhi\}_j$. Let $t_k=\theta_0<\theta_1<\dots<\theta_R=T_k$ collect these breakpoints within the window $[t_k,T_k]$, $T_k:=t_{k+1}+\taudhi$. On each slice $[\theta_r,\theta_{r+1}]$ the envelopes are constant, so $\vv^{\pm}$ are monotone there (Lemma~\ref{lem:flow}) and the reachable states lie in the slice box
\begin{equation}\label{eq:slicebox}
    \Bset_{k,r}=\big[\vv^{-}(\theta_r)\wedge\vv^{-}(\theta_{r+1}),\;\vv^{+}(\theta_r)\vee\vv^{+}(\theta_{r+1})\big],
\end{equation}
with $\wedge,\vee$ the componentwise minimum and maximum. The step-$k$ certificate, the lowest readiness the vehicle can reach over the window, is \begin{equation}\label{eq:cert}
    \cert(\tauv_k)=\min_{0\le r<R}\Lbox(\Bset_{k,r}).
\end{equation}
It needs the exact flows of Lemma~\ref{lem:flow} at the $R+1$ breakpoints and one $m\times m$ log-determinant per slice.

\begin{theorem}[Delay certificate]\label{thm:delay}
Let Assumption~\ref{ass:delay} hold and $\lbar>\ldrop$.
\begin{enumerate}
    \item[(i)] If $\cert(\tauv_k)\ge\lbar$, then $h(\vv(t))\ge 0$ for all $t\in[t_k,t_{k+1}]$,
    for every admissible realization of the switching instants and whatever commands are
    issued after $t_k$.
    \item[(ii)] If $\cert(\tauv_k)\ge\lbar$, then $\mathcal{L}_{k+1}(\tauh)\ge\lbar$.
    \item[(iii)] If $\mathcal{L}_0(\tauv_0)\ge\lbar$ and every later command either passes its certificate or equals $\tauh$, then $h(\vv(t))\ge 0$ for all $t\ge 0$; by (ii), a certified command exists at every step.
\end{enumerate}
\end{theorem}
\begin{proof}
(i) For $t<t_{k+1}$, $j\in\Iset(t)$ implies $t_j\le t-\taudlo<t_{k+1}$, i.e., $j\le k$: only committed commands and $\tauv_k$ act on $[t_k,t_{k+1}]$, so the hold hypothesis is immaterial there. By Lemma~\ref{lem:env} and the monotonicity of $\vv^{\pm}$ on each slice, $\vv(t)\in\Bset_{k,r}$ in \eqref{eq:slicebox} for $t\in[\theta_r,\theta_{r+1}]$. Since $\cert\ge\lbar>\ldrop$, Proposition~\ref{prop:boxmin} excludes $\underline v_i\le 0$ and gives $L(\vv(t))\ge\Lbox(\Bset_{k,r})\ge\lbar$.

(ii) With $\tauv_{k+1}=\tauh$, steps $k$ and $k+1$ use the same command sequence, so $\tauv^{\pm}(t)$ coincide for $t\ge t_{k+1}$. By Lemma~\ref{lem:env}, $\vv^{-}_k(t_{k+1})\le\vv_{k+1}\le\vv^{+}_k(t_{k+1})$, where $\vv^{\pm}_k$, $\vv^{\pm}_{k+1}$ are the envelopes of the two steps, and by Lemma~\ref{lem:flow} $\vv^{-}_k\le\vv^{-}_{k+1}\le\vv^{+}_{k+1}\le\vv^{+}_k$ for $t\ge t_{k+1}$. On $[t_{k+1},T_k]$ the breakpoints of step $k+1$ are those of step $k$ plus $t_{k+1}$, so every slice box of step $k+1$ lies in one of step $k$. On $[T_k,T_{k+1}]$, $\tauv^{\pm}\equiv\tauh$, under which each rotor moves monotonically toward $\vstar_i$ without crossing it, so $\psi_i(v^{\pm}_{k+1,i}(t))$ is nondecreasing (Lemma~\ref{lem:unimodal}) and every $\psilo_i$ there is at least $\psilo_i(\Bset_{k,R-1})$, which contains $[v^{-}_{k+1,i}(T_k),v^{+}_{k+1,i}(T_k)]$; Loewner monotonicity gives $\Lbox\ge\Lbox(\Bset_{k,R-1})\ge\lbar$.

(iii) Let $k'\le k$ be the last step whose command passed. All commands after $k'$ equal $\tauh$, the hypothesis of $\mathcal{L}_{k'}$, so the slice boxes of step $k'$ contain $\vv(t)$ on $[t_{k'},T_{k'}]$ (Lemma~\ref{lem:env}), and after $T_{k'}$ every rotor moves monotonically toward $\vstar_i$ from a point of $\Bset_{k',R-1}$, as in (ii). Hence $L(\vv(t))\ge\lbar$ until a new command passes; induction concludes.
\end{proof}

In practice, (i) says that an accepted command is safe until the next sample whatever the \acp{ESC} do, (ii) that the hold torque is always a certified fallback, and (iii) that the closed loop never leaves $\Cset$. Theorem~\ref{thm:delay} does not involve $\Lop$, but it needs a certified start: commands already in the pipeline may leave $\Cset$ before any new command acts.

\begin{corollary}[Source of conservatism]\label{cor:jitter}
For $t\in[t_k,T_k]$ and every $i$,
\begin{equation}\label{eq:spread}
    0\le v^{+}_i(t)-v^{-}_i(t)\le\frac{\jit}{m_i}\sum_{j\in\mathcal{K}_k(t)}|\tau_{j,i}-\tau_{j-1,i}|,
\end{equation}
where $\mathcal{K}_k(t)=\{j:[t_j+\taudlo,t_j+\taudhi)\cap[t_k,t]\neq\emptyset\}$ and $\tauv_j=\tauh$ for $j>k$. In particular, for a known constant delay ($\jit=0$), $\vv^{+}=\vv^{-}$ is the exact trajectory
under the hold hypothesis.
\end{corollary}
\begin{proof}
The spread $e_i=v^{+}_i-v^{-}_i$ starts at zero and, since $\phi$ is increasing, $m_i\dot e_i\le\tau^{+}_i-\tau^{-}_i$, which is at most the sum of $|\tau_{j,i}-\tau_{j-1,i}|$ over consecutive pairs in $\Iset(t)$. A pair $j-1,j$ is in $\Iset(t)$ exactly when $t\in[t_j+\taudlo,t_j+\taudhi)$, of length $\jit$; integrating gives \eqref{eq:spread}.
\end{proof}

Corollary~\ref{cor:jitter} explains when the certificate is conservative. The envelope spread, the part of the box width due to the uncertain switching, is $\jit$ times the total variation of the commands still in flight (about $\taudhi/\Delta t$ of them), the candidate, and the transition to the hold torque. It vanishes for a known delay and grows with the delay through the number of commands in flight, so smooth commands make the certificate tighter (Section~\ref{sec:sim}).



\subsection{Measurement error and drag uncertainty}
The certificate uses only monotonicity, so bounded model and measurement errors enter as wider envelopes. Suppose the filter measures $\hat{\vv}_k$ with $|\hat v_{k,i}-v_{k,i}|\le\epsm$ and the plant drag satisfies $b_i\in[b_i^{-},b_i^{+}]$. For $v_i>0$ the flow of \eqref{eq:rotor} is nonincreasing in $b_i$. Integrate $\vv^{+}$ from $\hat{\vv}_k+\epsm\mathbf{1}$ with $b_i^{-}$ and $\vv^{-}$ from $\hat{\vv}_k-\epsm\mathbf{1}$ with $b_i^{+}$. Under $\tauh$ the plant no longer settles at $\vstar$ but between $\veq_{i,-}=(\tau^{\mathrm{h}}_i/b_i^{+})^{1/2}$ and $\veq_{i,+}=(\tau^{\mathrm{h}}_i/b_i^{-})^{1/2}$, so we add the tail box $\Bset_{k,\infty}=[\vv^{-}(T_k)\wedge\vv^{\mathrm{eq}}_{-},\,\vv^{+}(T_k)\vee\vv^{\mathrm{eq}}_{+}]$ to \eqref{eq:cert}. In \eqref{eq:boxmin}, $\bar a_i$ is clipped at zero, so that $\psi_i$ stays unimodal on $[0,\infty)$ and boxes reaching beyond $\vsat_i$ are rejected.

\begin{proposition}[Robust certificate]\label{prop:robust}
Let $b_i^{-}>b_i/3$. With these envelopes and the tail box, Theorem~\ref{thm:delay}(i) and (iii) hold for every plant drag in $[\mathbf{b}^{-},\mathbf{b}^{+}]$ and every measurement error bounded by $\epsm$. If $\epsm=0$, Theorem~\ref{thm:delay}(ii) holds as well.
\end{proposition}
\begin{proof}
A passing certificate has $\vv^{-}>\mathbf{0}$ on the window. The true state starts in $[\hat{\vv}_k-\epsm\mathbf{1},\hat{\vv}_k+\epsm\mathbf{1}]$ and, for positive speeds, the right-hand side of \eqref{eq:rotor} is nonincreasing in $b_i$ and nondecreasing in $\tau_i$, so the comparison principle brackets it between $\vv^{\pm}$: this gives (i). After $T_k$, each true rotor moves monotonically from $[v_i^{-}(T_k),v_i^{+}(T_k)]$ toward its equilibrium in $[\veq_{i,-},\veq_{i,+}]$, so it stays in $\Bset_{k,\infty}$: this gives (iii). For (ii) with $\epsm=0$, the slices on $[t_{k+1},T_k]$ nest as in Theorem~\ref{thm:delay}, and the step-$(k+1)$ tail box lies in $\Bset_{k,\infty}$.
\end{proof}

The certificate can pass only if $\Lbox([\vv^{\mathrm{eq}}_{-},\vv^{\mathrm{eq}}_{+}])\ge\lbar$ (which needs $b_i^{-}>b_i/3$); for the hexarotor with $\pm10\%$ drag this value is $-10.16$, far above the floors of Section~\ref{sec:sim}. Two remarks delimit the result. First, the certified floor is the nominal readiness \eqref{eq:readiness} of the true speeds, not that of the perturbed plant (reported in Section~\ref{sec:sim}). Second, with $\epsm>0$ the re-measured box at $t_{k+1}$ may exceed the step-$k$ envelopes by $2\epsm$, so a fresh certificate may fail; applying $\tauh$ then continues the last certified plan, which remains safe by (iii).

\emph{Inner \ac{ESC} loops.} Lemma~\ref{lem:env} needs only flows monotone in the initial state and the command. A proportional speed loop, $\tau_i=\mathrm{sat}(k_p(v^{\mathrm{c}}_i-v_i))$, keeps each rotor scalar and monotone in the speed command $v^{\mathrm{c}}_i$. A first-order torque lag, $T_{\mathrm{e}}\dot\tau^{\mathrm{a}}_i=\tau_i-\tau^{\mathrm{a}}_i$, makes $(v_i,\tau^{\mathrm{a}}_i)$ cooperative, so Lemma~\ref{lem:env} holds on the augmented state with numerical envelopes~\cite{angeli2003monotone}, but the tail of Theorem~\ref{thm:delay}(ii) needs an explicit bound, because the speed may first move away from $\vstar_i$. Integral action breaks cooperativity and needs a mixed-monotone embedding~\cite{coogan2020mixed}.



\subsection{Certified allocation filter}
Algorithm~\ref{alg:filter} wraps the \ac{QP}~\eqref{eq:qp}. The \ac{QP} is solved at the state predicted at $t_k+\taudnom$, with $\taudnom=(\taudlo+\taudhi)/2$, by integrating the committed commands with the exact flows. Its output is applied if it passes the certificate. Otherwise the command is blended toward $\tauh$, which passes by Theorem~\ref{thm:delay}(ii) on the exact model. Bisection on the blend factor $\gamma\in[0,1]$ keeps only certified points (the smallest certified $\gamma$ is not guaranteed: blending may enlarge the increment $\tauv_k-\tauv_{k-1}$ in \eqref{eq:spread}). If the \ac{QP} fails (e.g., an infeasible barrier row at a noisy measurement) or $\tauh$ fails a fresh check, $\tauh$ is applied, which continues the last certified plan (Theorem~\ref{thm:delay}(iii), Proposition~\ref{prop:robust}). The fallback is task-agnostic and perturbs the wrench, in particular the collective.

\begin{algorithm}[t]
    \caption{Delay-certified readiness filter (step $k$)}\label{alg:filter}
    \begin{algorithmic}[1]
    \REQUIRE $\vv_k$, committed $\{\tauv_j\}_{j<k}$, $[\taudlo,\taudhi]$, $\lbar$, $N_b$; a passing
    certificate at some earlier step
    \STATE $\hat{\vv}\leftarrow$ exact flow of $\vv_k$ to $t_k+\taudnom$ under committed commands
    \STATE $\tauv_{\mathrm{ref}}\leftarrow$ effort-minimizing allocation at $\hat{\vv}$;
    $\tauv^{\mathrm{QP}}\leftarrow$ QP~\eqref{eq:qp} at $\hat{\vv}$
    \IF{the QP is solved and $\cert(\tauv^{\mathrm{QP}})\ge\lbar$} \RETURN $\tauv^{\mathrm{QP}}$ \ENDIF
    \IF{$\cert(\tauh)<\lbar$ or the QP failed} \RETURN $\tauh$ \COMMENT{retained plan} \ENDIF
    \STATE $\gamma^{-}\leftarrow 0$, $\gamma^{+}\leftarrow 1$
    \FOR{$N_b$ iterations}
    \STATE $\gamma\leftarrow(\gamma^{-}+\gamma^{+})/2$
    \IF{$\cert((1-\gamma)\tauv^{\mathrm{QP}}+\gamma\tauh)\ge\lbar$} \STATE $\gamma^{+}\leftarrow\gamma$ \ELSE \STATE $\gamma^{-}\leftarrow\gamma$ \ENDIF
    \ENDFOR
    \RETURN $(1-\gamma^{+})\tauv^{\mathrm{QP}}+\gamma^{+}\tauh$
    \end{algorithmic}
\end{algorithm}



\section{Simulation Study}\label{sec:sim}

The simulations answer four questions:
\begin{itemize}
    \item[\textbf{Q1}] How tight is the box bound? 
    \item[\textbf{Q2}] Does the filter keep the floor across delays and jitter, and at what cost (Sections~\ref{sec:sweep} and~\ref{sec:mc})? 
    \item[\textbf{Q3}] How does it compare with Lipschitz margins (Section~\ref{sec:mc})?
    \item[\textbf{Q4}] Does it survive model and measurement errors and \ac{ESC} timing and dynamics (Sections~\ref{sec:mismatch} and~\ref{sec:async})?
\end{itemize}

\emph{Platform and missions.} We use the coplanar hexarotor of~\cite{silano2026readiness}, overactuated with respect to its $m=4$ controlled wrench components (collective thrust and three moments): $n=6$ identical nondimensional motors $\taubar_i=b_i=m_i=1$, with $T_0=0.32$~s, and $\Am_{\bullet i}=(1,\,l\sin\beta_i,\,-l\cos\beta_i,\, \kappa_{\mathrm{y}}(-1)^{i+1})$ with $\beta_i=30^\circ+60^\circ(i-1)$, $l=0.25$, and $\kappa_{\mathrm{y}}=0.016$, so that $\Lmax=-10.128$ and $\ldrop=-11.226$. The mission is the roll-moment reversal of~\cite{silano2026readiness}: fixed collective $c$ (fraction of the sweet-spot collective) and a sinusoidal roll moment of amplitude $\beta$ (in units of $\phi(\vstar_i)/4$) and period $2T_0$, flown for $2T_0$. On \emph{certifiable} missions the demand stays within the design envelope ($\Lop>\ldrop$, computed on 41 points of one period) and the floor follows the rule of~\cite{silano2026readiness}, $\lbar=\ldrop+\frac12(\Lop-\ldrop)$. Since floors are fixed at design while maneuvers are commanded online, we also fly \emph{readiness-priority} missions, whose demand exceeds the envelope: the floor $\lbar=\ldrop+0.15>\Lop$ cannot be held while tracking exactly~\cite[Prop.~2]{silano2026readiness}, so every filter must trade tracking for readiness.

\emph{Implementation.} The filter runs at $\Delta t=2$~ms; the plant is integrated piecewise between switching instants (fourth-order Runge--Kutta, steps $\le10^{-4}T_0$). Each command switches at $\sigma_{j,i}=\sigma_j=\max\{t_j+\taudnom+\epsilon_j,\sigma_{j-1}\}$, with $\epsilon_j$ uniform in $[-\jit/2,\jit/2]$ (per-rotor instants in Section~\ref{sec:async}). Runs start at the readiness-maximizing allocation of $\wv_{\mathrm{des}}(0)$ with the queue at its equilibrium torque, and the first certificate passed in every run. All filters use the gains of~\cite{silano2026readiness} ($\alpha(h)=\alpha h$, $\alpha=5$ unless stated, $k_w=8$, $\rho=10^4$), OSQP~\cite{stellato2020osqp}, and the same missions and jitter realizations (fixed seeds); Algorithm~\ref{alg:filter} uses $N_b=14$. 

\emph{Baselines.} All are built on~\eqref{eq:qp}. The \emph{RB filter}, i.e., \eqref{eq:qp} at the measured state~\cite{silano2026readiness}, shows what happens if the delay is ignored. \emph{Pred}, i.e., \eqref{eq:qp} at the state predicted at $t_k+\taudnom$ with exact flows (predictor feedback~\cite{molnar2023delay,kiss2026icbf} without robustness margin), isolates delay compensation. The others add the conventional safety margin $\zeta$ to the barrier row. \emph{SD-Lip}: the RB filter with $\zeta=KV\Delta t$, as in the global conditions of~\cite{breeden2022sampled}. \emph{Pred-Lip}: Pred with $\zeta=KV(\Delta t+\jit/2)$, a heuristic allowance for the hold interval and half the jitter (it does not bound the prediction error, which accumulates as in~\eqref{eq:spread}). For these two, $K$ (a bound on $\|\nabla(\nabla h^{\top}\dv)\|_1+\|\nabla^2h\|_1\max_i\taubar_i/m_i$) and $V$ (on $\|\dot{\vv}\|_\infty$) are sampled on $\Cset$: $KV=212\,T_0^{-2}$, about half of the global $398\,T_0^{-2}$ of~\cite{silano2026readiness}. \emph{Pred-Lip-G}: the same margin with the global constants. \emph{Pred-Lip-Loc}: the same margin with $K_kV_k$ sampled at every step at $\vv_k$ and 32 random points of the box of radius $V_{\max}(\Delta t+\taudhi)$ around it, $V_{\max}=\max_i2\taubar_i/m_i$, which contains the speeds reachable over the window. None of these margins is a certificate; when the barrier row is infeasible, they apply the torque that maximizes $\dot h$. \emph{Pred-ISSf}: Pred with the input-to-state-safety term $\|\nabla h^{\top}\Mm^{-1}\|^2/\varepsilon$ on the barrier row, as in the robust predictor-feedback \acp{CBF} of~\cite{molnar2022issf, molnar2023delay}; against the input disturbance $\mathbf{e}$ caused by the prediction error, it guarantees only $h\ge-\varepsilon\|\mathbf{e}\|_\infty^2/(4\alpha)$, so certifying the floor would again need a bound on $\mathbf{e}$. We sweep $\varepsilon\in\{10,30,100\}$ and report $\varepsilon=30$, tuned a posteriori in its favor.

\emph{Metrics.} The worst margin $\min_t h$ over a run, recorded at every integration step, is what the filter guarantees: a negative value means that the readiness fell below the floor. The \ac{RMS} wrench error (Euclidean norm, nondimensional) relative to Pred, $e_w$ (median over missions of the paired ratio), isolates the price of the extra safety, because every filter enforces the floor by giving up tracking and Pred is the cheapest filter that compensates the delay. The fallback rate shows how often the certificate intervenes.

\textbf{Q1}--\emph{Tightness.} Over one hold interval without delay, on 400 random states, the gap between the box bound and the true minimum grows linearly in $\Delta t$, that of a second-order Lipschitz bound with the constant sampled on $\Cset$ quadratically: the Lipschitz bound is tighter at 0.5~ms, the box bound 1.9 and 11 times tighter at 2 and 10~ms.



\subsection{Delay and jitter sweep}\label{sec:sweep}
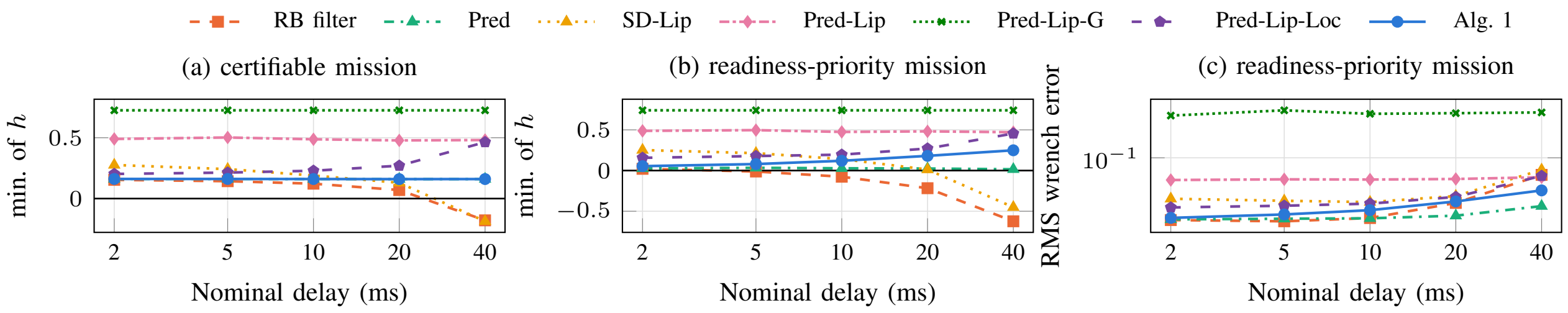
\begin{figure*}[t]
    \centering
\definecolor{cB1}{RGB}{235,104,52}
\definecolor{cB1p}{RGB}{27,175,122}
\definecolor{cB3}{RGB}{237,161,0}
\definecolor{cB4}{RGB}{232,123,164}
\definecolor{cB4g}{RGB}{0,131,0}
\definecolor{cB4l}{RGB}{118,68,160}
\definecolor{cP}{RGB}{42,120,214}
\begin{tikzpicture}
\begin{groupplot}[group style={group size=3 by 1, horizontal sep=1.25cm},
  width=0.335\textwidth, height=3.0cm, xmode=log, log basis x=10,
  xtick={2,5,10,20,40}, xticklabels={2,5,10,20,40}, minor xtick={}, xmin=1.7, xmax=47,
  xlabel={Nominal delay (ms)}, label style={font=\footnotesize}, tick label style={font=\scriptsize},
  title style={font=\footnotesize, yshift=-0.8ex}, every axis plot/.append style={line width=0.8pt, mark size=1.4pt},
  legend columns=7, legend style={font=\scriptsize, draw=none, column sep=0.6em,
    at={(1.85,1.40)}, anchor=south}]
\nextgroupplot[title={(a) certifiable mission}, ylabel={min.\ of $h$}]
\addplot[color=cB1, dashed, mark=square*, mark options={solid, fill=cB1}] coordinates {(2,0.153828) (5,0.142834) (10,0.122362) (20,0.0699348) (40,-0.178954)};
\addlegendentry{RB filter}
\addplot[color=cB1p, dashdotted, mark=triangle*, mark options={solid, fill=cB1p}] coordinates {(2,0.161182) (5,0.161208) (10,0.160521) (20,0.159901) (40,0.160292)};
\addlegendentry{Pred}
\addplot[color=cB3, dotted, mark=triangle*, mark options={solid, fill=cB3}] coordinates {(2,0.276081) (5,0.242039) (10,0.189436) (20,0.129772) (40,-0.186651)};
\addlegendentry{SD-Lip}
\addplot[color=cB4, dash pattern=on 3pt off 1pt on 1pt off 1pt, mark=diamond*, mark options={solid, fill=cB4}] coordinates {(2,0.489793) (5,0.501995) (10,0.487068) (20,0.478021) (40,0.481025)};
\addlegendentry{Pred-Lip}
\addplot[color=cB4g, densely dotted, mark=x, mark options={solid, fill=cB4g}] coordinates {(2,0.725157) (5,0.725157) (10,0.725157) (20,0.725157) (40,0.725157)};
\addlegendentry{Pred-Lip-G}
\addplot[color=cB4l, loosely dashed, mark=pentagon*, mark options={solid, fill=cB4l}] coordinates {(2,0.202386) (5,0.213571) (10,0.228698) (20,0.270078) (40,0.4635)};
\addlegendentry{Pred-Lip-Loc}
\addplot[color=cP, solid, mark=*, mark options={solid, fill=cP}] coordinates {(2,0.161182) (5,0.161208) (10,0.160521) (20,0.159901) (40,0.160292)};
\addlegendentry{Alg.~1}
\addplot[black, line width=0.5pt, forget plot, mark=none] coordinates {(1.7,0) (47,0)};
\nextgroupplot[title={(b) readiness-priority mission}, ylabel={min.\ of $h$}]
\addplot[color=cB1, dashed, mark=square*, mark options={solid, fill=cB1}] coordinates {(2,0.0212138) (5,-0.011778) (10,-0.0758365) (20,-0.216954) (40,-0.623513)};
\addplot[color=cB1p, dashdotted, mark=triangle*, mark options={solid, fill=cB1p}] coordinates {(2,0.0256117) (5,0.0334426) (10,0.0277458) (20,0.0247134) (40,0.0154008)};
\addplot[color=cB3, dotted, mark=triangle*, mark options={solid, fill=cB3}] coordinates {(2,0.251944) (5,0.213528) (10,0.140892) (20,0.0124894) (40,-0.450681)};
\addplot[color=cB4, dash pattern=on 3pt off 1pt on 1pt off 1pt, mark=diamond*, mark options={solid, fill=cB4}] coordinates {(2,0.487613) (5,0.496624) (10,0.474581) (20,0.481446) (40,0.47112)};
\addplot[color=cB4g, densely dotted, mark=x, mark options={solid, fill=cB4g}] coordinates {(2,0.740148) (5,0.740148) (10,0.740148) (20,0.740148) (40,0.740148)};
\addplot[color=cB4l, loosely dashed, mark=pentagon*, mark options={solid, fill=cB4l}] coordinates {(2,0.15646) (5,0.17818) (10,0.195248) (20,0.270596) (40,0.454869)};
\addplot[color=cP, solid, mark=*, mark options={solid, fill=cP}] coordinates {(2,0.0539477) (5,0.0796662) (10,0.119818) (20,0.180444) (40,0.248908)};
\addplot[black, line width=0.5pt, forget plot, mark=none] coordinates {(1.7,0) (47,0)};
\nextgroupplot[title={(c) readiness-priority mission}, ylabel={RMS wrench error}, ymode=log]
\addplot[color=cB1, dashed, mark=square*, mark options={solid, fill=cB1}] coordinates {(2,0.0156164) (5,0.0150502) (10,0.016584) (20,0.0260306) (40,0.0601135)};
\addplot[color=cB1p, dashdotted, mark=triangle*, mark options={solid, fill=cB1p}] coordinates {(2,0.0159446) (5,0.0162448) (10,0.0164559) (20,0.0177942) (40,0.0236567)};
\addplot[color=cB3, dotted, mark=triangle*, mark options={solid, fill=cB3}] coordinates {(2,0.0294875) (5,0.0277574) (10,0.0266798) (20,0.0319624) (40,0.0698273)};
\addplot[color=cB4, dash pattern=on 3pt off 1pt on 1pt off 1pt, mark=diamond*, mark options={solid, fill=cB4}] coordinates {(2,0.0515482) (5,0.0526267) (10,0.0522718) (20,0.0531635) (40,0.0559933)};
\addplot[color=cB4g, densely dotted, mark=x, mark options={solid, fill=cB4g}] coordinates {(2,0.352439) (5,0.412339) (10,0.371594) (20,0.378604) (40,0.387039)};
\addplot[color=cB4l, loosely dashed, mark=pentagon*, mark options={solid, fill=cB4l}] coordinates {(2,0.0227408) (5,0.0239431) (10,0.0254944) (20,0.0311075) (40,0.0583278)};
\addplot[color=cP, solid, mark=*, mark options={solid, fill=cP}] coordinates {(2,0.01666) (5,0.0184534) (10,0.0209697) (20,0.0272158) (40,0.0378221)};
\end{groupplot}
\end{tikzpicture}
    \caption{Nominal-delay sweep at jitter $\pm2$~ms ($\jit=4$~ms), $\alpha=5$, synchronous switching, five jitter realizations per point; $c=0.75$, (a) $\beta=1.5$, $\lbar=-11.061$, (b), (c) $\beta=2$, $\lbar=-11.076$. (a), (b): worst $\min_t h$ over the realizations (below zero: floor violated); (c): median RMS wrench error. Algorithm~1 stays above the floor at every delay, while the RB filter and SD-Lip fail at long delays; in (a), Algorithm~1 and Pred coincide.}
    \label{fig:delaymap}
\end{figure*}

\textbf{Q2}--Algorithm~\ref{alg:filter} keeps the floor at every delay and jitter tested, whereas ignoring the delay fails as the delay grows. Figure~\ref{fig:delaymap} sweeps the nominal delay from 2 to 40~ms on two missions with $c=0.75$: a certifiable one ($\beta=1.5$, $\lbar=-11.061$) and a readiness-priority one ($\beta=2.0$, $\Lop=-11.405$, $\lbar=-11.076$). Each of 20 delay--jitter configurations ($\taudnom\in\{2,5,10,20,40\}$~ms, $\jit\in\{0,1,2,4\}$~ms) is flown on both missions with five jitter realizations: Algorithm~\ref{alg:filter} keeps the floor in all 200 runs.

On the certifiable mission, Algorithm~\ref{alg:filter} coincides with Pred in all 100 runs, while the RB filter and SD-Lip violate at 40~ms ($\min h=-0.18$ and $-0.19$), consistent with the failure near 29~ms reported in~\cite{silano2026readiness}. On the readiness-priority mission, the RB filter violates from 5~ms ($-0.62$ at 40~ms); Pred does not violate at $\alpha=5$ ($\min h\ge0.015$), Pred-Lip keeps $\min h\ge0.27$ at 1.5--3.3 times its wrench error, and Algorithm~\ref{alg:filter} $\min h\ge0.038$ at 1.0--1.65 times. Pred-Lip-Loc never violates ($\min h\ge0.105$), at 1.1--5.3 times the wrench error of Pred, more than Algorithm~\ref{alg:filter} at every delay in Fig.~\ref{fig:delaymap}(c), as in Table~\ref{tab:mc} at $\alpha=5$. The fallback of Algorithm~\ref{alg:filter} is never active without jitter, as Corollary~\ref{cor:jitter} suggests, and at $\jit=4$~ms it is active in 11\% of the steps at 2~ms and 35\% at 40~ms, because a longer window holds more commands in flight.

\subsection{Monte Carlo missions}\label{sec:mc}
\textbf{Q2}--Beyond the design envelope, delay compensation alone is not enough, and the certificate keeps the floor at a moderate cost. Table~\ref{tab:mc} reports 50 missions with $c\sim\mathcal{U}[0.7,0.9]$ and $\beta\sim\mathcal{U}[1.0,2.5]$, flown at $10\pm2$~ms with $\alpha=5$ and $\alpha=20$; 34 are certifiable and 16 are flown with the readiness-priority floor. On the exact model, the zero violations of Algorithm~\ref{alg:filter} confirm the implementation of Theorem~\ref{thm:delay}. On certifiable missions its median wrench error equals that of Pred (fallback in 2\% and 0.1\% of the steps; worst mission at $\alpha=5$: 1.6 times Pred). Beyond the envelope the predictor's safety depends on the gain: Pred violates in 2 of 16 missions at $\alpha=5$ and in all 16 at $\alpha=20$, while Algorithm~\ref{alg:filter} holds the floor at a median wrench-error increase of 19\% and 30\%. With a thin buffer, $\lbar=\ldrop+0.9(\Lop-\ldrop)$, on certifiable missions, Pred never violates: the certificate adds a guarantee, not observed safety.

\textbf{Q3}--The Lipschitz margins are also safe, without a guarantee and at a cost that depends strongly on the gain and on the constants. Pred-Lip costs 7.7 and 2.4 times the median wrench error of Pred at $\alpha=5$, but 1.00 and 1.35 times at $\alpha=20$: near $\vstar$, where $\nabla h\approx\mathbf{0}$ and $h\le\Lmax-\lbar$, the barrier row is feasible only if $\alpha(\Lmax-\lbar)\ge\zeta$, with $\alpha(\Lmax-\lbar)\in[3.6,5.4]$ at $\alpha=5$ and $[14,22]$ at $\alpha=20$, against $\zeta=2.65$ (Pred-Lip) and $4.97\,T_0^{-1}$ (Pred-Lip-G, omitted from Table~\ref{tab:mc}: at $\alpha=5$, infeasible in 60\% of the steps and 142 times Pred; at $\alpha=20$, 1.1--1.7 times). Pred-Lip-Loc costs 1.21 and 1.45 times Pred at $\alpha=5$ and 1.00 and 1.19 at $\alpha=20$: at the high gain, a local margin is cheaper than the certificate beyond the envelope. Pred-ISSf ($\varepsilon=30$) is safe in all 100 runs, at 1.19 and 1.09 times Pred beyond the envelope, but its tuning is fragile: $\varepsilon=100$ violates in 15 of 16 such missions at $\alpha=20$, and $\varepsilon=10$ costs 1.53 and 1.25 times Pred.

\begin{table*}[t]
\centering
    \caption{Monte Carlo missions at $10\pm2$~ms. Viol.: missions with $\min_t h<0$ (bold);
    $h_{\min}$: worst $\min_t h$; $e_w$: median (90th percentile) of the RMS wrench
    error relative to Pred (absolute medians, sweet-spot collective $=2$: 0.0023, 0.0022
    certifiable; 0.0257, 0.0202 readiness-priority; $\alpha=5$, 20); act.: \% of steps with
    active fallback (Algorithm~1). Algorithm~1 never violates and matches Pred on certifiable missions; beyond the envelope, Pred violates and the others are safe but, except Pred-Lip-Loc and the tuned Pred-ISSf at $\alpha=20$, costlier.}
    \label{tab:mc}
    \begin{tabular}{lrrrrrrrr}
\toprule
& \multicolumn{4}{c}{$\alpha=5$} & \multicolumn{4}{c}{$\alpha=20$}\\
\cmidrule(lr){2-5}\cmidrule(lr){6-9}
Filter & viol. & $h_{\min}$ & $e_w$ & act. & viol. & $h_{\min}$ & $e_w$ & act.\\
\midrule
\multicolumn{9}{l}{\emph{certifiable missions (34)}}\\
RB filter & 0 & $+0.014$ & 2.35 (3.05) & 0 & \textbf{3} & $\mathbf{-0.035}$ & 2.72 (3.36) & 0 \\
Pred & 0 & $+0.098$ & 1.00 (1.00) & 0 & 0 & $+0.013$ & 1.00 (1.00) & 0 \\
Pred-ISSf & 0 & $+0.158$ & 1.00 (1.67) & 0 & 0 & $+0.035$ & 1.00 (1.00) & 0 \\
Pred-Lip & 0 & $+0.408$ & 7.70 (12.29) & 0 & 0 & $+0.135$ & 1.00 (1.46) & 0 \\
Pred-Lip-Loc & 0 & $+0.204$ & 1.21 (2.77) & 0 & 0 & $+0.058$ & 1.00 (1.00) & 0 \\
Alg.~1 & 0 & $+0.130$ & 1.00 (1.20) & 2.0 & 0 & $+0.015$ & 1.00 (1.00) & 0.1 \\
\midrule
\multicolumn{9}{l}{\emph{readiness-priority missions (16)}}\\
RB filter & \textbf{16} & $\mathbf{-0.111}$ & 0.94 (1.07) & 0 & \textbf{16} & $\mathbf{-0.106}$ & 1.15 (1.45) & 0 \\
Pred & \textbf{2} & $\mathbf{-0.004}$ & 1.00 (1.00) & 0 & \textbf{16} & $\mathbf{-0.022}$ & 1.00 (1.00) & 0 \\
Pred-ISSf & 0 & $+0.095$ & 1.19 (1.35) & 0 & 0 & $+0.011$ & 1.09 (1.17) & 0 \\
Pred-Lip & 0 & $+0.480$ & 2.39 (3.78) & 0 & 0 & $+0.109$ & 1.35 (1.78) & 0 \\
Pred-Lip-Loc & 0 & $+0.156$ & 1.45 (1.70) & 0 & 0 & $+0.026$ & 1.19 (1.32) & 0 \\
Alg.~1 & 0 & $+0.121$ & 1.19 (1.35) & 30 & 0 & $+0.023$ & 1.30 (1.59) & 32 \\
\bottomrule
\end{tabular}

\end{table*}



\subsection{Model mismatch and measurement error}\label{sec:mismatch}
\textbf{Q4}--With drag and measurement errors, the robust certificate still keeps the floor, and also the readiness of the true plant. Table~\ref{tab:mismatch} repeats the $\alpha=5$ Monte Carlo with a plant whose drag coefficients deviate by up to $\pm10\%$ per rotor and with speed measurements corrupted by uniform errors of up to $\epsm=0.005\,\vsat$. Algorithm~1-R uses the robust certificate of Proposition~\ref{prop:robust} with the true bounds. Pred violates in 3 of 16 readiness-priority missions, neither variant of Algorithm~\ref{alg:filter} in any. As Proposition~\ref{prop:robust} anticipates, the hold torque fails a fresh check in 0.8\% of the steps and is then applied as the retained plan. Evaluated with the plant drag, only Algorithm~1-R keeps the readiness above the floor ($h^{\mathrm{true}}_{\min}\ge+0.026$; Pred and the nominal Algorithm~1: $-0.135$ and $-0.028$).

\begin{table}[t]
    \centering
    \caption{As Table~\ref{tab:mc} with $\pm10\%$ drag mismatch and $0.5\%$ speed error, $\alpha=5$.
    $h^{\mathrm{true}}_{\min}$: with the plant drag; unc.: runs (\% of steps) with a failed fresh
    hold check. Absolute $e_w$ of Pred: 0.0062, 0.0266. Only Algorithm~1-R keeps the true-plant readiness above the floor.}
    \label{tab:mismatch}
    \setlength{\tabcolsep}{4pt}
    \begin{tabular}{lrrrrrr}
\toprule
Filter & viol. & $h_{\min}$ & $h^{\mathrm{true}}_{\min}$ & $e_w$ & act. & unc.\\
\midrule
\multicolumn{7}{l}{\emph{certifiable missions (34)}}\\
Pred & 0 & $+0.090$ & $\mathbf{-0.018}$ & 1.00 (1.00) & 0 & -- \\
Pred-Lip & 0 & $+0.399$ & $+0.224$ & 3.61 (6.15) & 0 & -- \\
Alg.~1 & 0 & $+0.124$ & $\mathbf{-0.004}$ & 1.00 (1.13) & 1.8 & 6 (0.1) \\
Alg.~1-R & 0 & $+0.147$ & $+0.026$ & 1.00 (1.27) & 3.3 & 8 (0.1) \\
\midrule
\multicolumn{7}{l}{\emph{readiness-priority missions (16)}}\\
Pred & \textbf{3} & $\mathbf{-0.007}$ & $\mathbf{-0.135}$ & 1.00 (1.00) & 0 & -- \\
Pred-Lip & 0 & $+0.471$ & $+0.370$ & 2.32 (3.34) & 0 & -- \\
Alg.~1 & 0 & $+0.121$ & $\mathbf{-0.028}$ & 1.22 (1.38) & 27 & 16 (1.8) \\
Alg.~1-R & 0 & $+0.155$ & $+0.033$ & 1.31 (1.54) & 32 & 16 (2.1) \\
\bottomrule
\end{tabular}

\end{table}



\emph{Source of the cost.} On the readiness-priority mission of Fig.~\ref{fig:delaymap} at $10\pm2$~ms, the extra wrench error of Algorithm~\ref{alg:filter} (0.0209 against 0.0165 for Pred; fallback in 27\% of the steps) is mostly in the collective, which the hold torque perturbs (0.0088 against 0.0046). As Corollary~\ref{cor:jitter} suggests, penalizing $10\|\tauv-\tauv_{k-1}\|^2$ in the QP halves the fallback rate and lowers the error to $0.0188$.



\subsection{Independent \ac{ESC} jitter and lag}\label{sec:async}
\textbf{Q4}--Real \acp{ESC} switch at slightly different instants. With independent per-rotor offsets $\epsilon_{j,i}$ in the Monte Carlo of Table~\ref{tab:mc}, Pred still holds the floor on certifiable missions, but beyond the envelope it violates in 11 of 16 missions at $\alpha=5$ (2 with synchronous switching; $\min h$ down to $-0.27$) and in all 16 at $\alpha=20$. Algorithm~\ref{alg:filter} keeps the floor in all 100 runs at 1.09 and 1.02 times the wrench error of Pred, because Lemma~\ref{lem:env} covers the skew by construction. Pred-ISSf, with the $\varepsilon$ safe under synchronous switching, violates in 4 and 16 of 16 missions beyond the envelope ($\min h$ down to $-0.18$) and in 2 certifiable missions at $\alpha=20$.

\emph{Unmodeled \ac{ESC} lag.} In the $\alpha=5$ Monte Carlo of Table~\ref{tab:mc}, each rotor now tracks its delayed torque through a first-order lag ($T_{\mathrm{e}}=5$ or 20~ms) unknown to every filter. Pred violates in 7 of 16 readiness-priority missions at 5~ms, and at 20~ms in 7 of 34 certifiable and all 16 readiness-priority ones ($\min h$ down to $-0.13$). Algorithm~\ref{alg:filter} and Pred-ISSf keep the floor in all 100 runs, beyond the envelope at 1.21 and 1.19 times Pred at 5~ms and 1.25 and 1.08 at 20~ms, where the margins are thin ($+0.013$, $+0.004$) and the fresh hold check of Algorithm~\ref{alg:filter} fails in 15 of 16 runs: safety is observed, not certified.



\subsection{Computation}
The certificate is cheap: at $10\pm2$~ms it covers 7 slices, each with $2n$ closed-form flows and one $4\times4$ Cholesky factorization, some $3\times10^3$ operations in all. Written in Python and compiled to machine code with Numba, a just-in-time compiler for numerical Python, it takes $9.6~\mu$s (median; 99th percentile $14.6~\mu$s) on one core of an Intel Core Ultra 7 255H CPU under Ubuntu 24.04. A full filter step (prediction, two OSQP solves, certificates) takes 0.57~ms (median; maximum 1.42~ms) with $N_b=14$, against 0.48~ms for Pred. These are not hard bounds; on flight hardware, the computation time must be measured and included in $\taudlo$.



\section{Conclusion}\label{sec:concl}

Readiness barriers can be certified under sampling and uncertain actuation delay without Lipschitz constants: monotone closed-form rotor flows and a readiness unimodal in each rotor speed make the worst case over all reachable speeds a few closed-form evaluations, and the sweet-spot hold torque provides a certified fallback.

For a practitioner, the certificate turns a delay known only as an interval into a guarantee: it wraps any allocation \ac{QP}, takes microseconds, and matched the tracking error of an uncertified predictor when the floor was compatible with the mission. It pays off when demands may exceed the design envelope or \acp{ESC} switch at different instants, where the predictor alone violated the floor; otherwise the predictor was safe, and local Lipschitz margins or a tuned robust predictor can be cheaper at high gains, without a guarantee and, for the latter, without robustness to per-rotor jitter.

\emph{Limitations.} The guarantee needs the rotor model of~\cite{silano2026readiness}, a certified start, hard delay bounds including computation, \acp{ESC} without inner dynamics, no reordering or dropped packets, and exact arithmetic; with measurement error, only the last certified plan is guaranteed, on the nominal readiness. The evidence is simulation of one hexarotor. Next steps are flight tests with measured \ac{ESC} delays, a wrench-preserving backup, and the articulated platform of~\cite{silano2026tilt}.



\bibliographystyle{IEEEtran}
\bibliography{refs}

\end{document}